%% file: main.tex
\documentclass[sigconf,nonacm]{acmart}

\setcopyright{none}
\renewcommand\footnotetextcopyrightpermission[1]{}

\usepackage{booktabs}
\usepackage{graphicx}
\usepackage{float}
\usepackage{amsmath}
\usepackage{xcolor}
\newcommand{\OpusVerifier}{Opus~4.8}
\newcommand{\JevVerifier}{TypeSafe Jev}
\newcommand{\Metric}[1]{\csname result#1\endcsname}
\input{generated/metrics.tex}

\begin{document}

\title{Reward Integrity in Adversarial Reinforcement Learning for LLM Guardrail Red-Teaming}

\author{Chengming Liu}
\affiliation{%
  \institution{Commonwealth Bank of Australia}
  \city{Sydney}
  \country{Australia}}
\email{Ray.Liu@cba.com.au}

\author{Leon Viti}
\affiliation{%
  \institution{Commonwealth Bank of Australia}
  \city{Sydney}
  \country{Australia}}
\email{Leon.Viti@cba.com.au}

\author{Hongxu Chen}
\affiliation{%
  \institution{Commonwealth Bank of Australia}
  \city{Sydney}
  \country{Australia}}
\email{Hongxu.Chen@cba.com.au}

\author{Erdem Kemik}
\affiliation{%
  \institution{Commonwealth Bank of Australia}
  \city{Sydney}
  \country{Australia}}
\email{Erdem.Kemik@cba.com.au}

\author{Mary-Anne Williams}
\affiliation{%
  \institution{Commonwealth Bank of Australia}
  \city{Sydney}
  \country{Australia}}
\email{Mary-Anne.Williams@cba.com.au}

\author{Ruth Oliver}
\affiliation{%
  \institution{Commonwealth Bank of Australia}
  \city{Sydney}
  \country{Australia}}
\email{Ruth.Oliver@cba.com.au}

\renewcommand{\shortauthors}{Liu et al.}

\begin{abstract}
We study reward fidelity in an adversarial reinforcement-learning loop for
red-teaming financial-services language-model guardrails. Apparent wins can
arise when an attack omits prescribed personally identifiable information
(PII), or when a reward judge credits a response that did not break the targeted
policy. On attacker-generated and public benchmark examples, TypeSafe Jev
checks whether the prescribed PII is present. For outcome scoring, we compare
the decision models TypeSafe Jev, Laya (pretrained and post-trained), and
pretrained CLM-8B with logged Opus~4.8 judgements from a replay corpus.
Post-training markedly improves Laya's agreement with the judge. A separate
author-reviewed off-target set establishes that Opus can award false wins.
The findings motivate specialised decision-model judges for checking prescribed
PII in attack inputs and scoring target responses against the intended policy.
\end{abstract}


\maketitle

\input{sections/01-introduction}
\input{sections/02-loop}
\input{sections/03-signal-fidelity}
\input{sections/04-failure-modes}
\input{sections/05-results}
\input{sections/06-related-conclusion}

\bibliographystyle{ACM-Reference-Format}
\bibliography{references}

\end{document}

%% file: generated/metrics.tex
\expandafter\def\csname resultattackerArchivedBA\endcsname{87.8}
\expandafter\def\csname resultattackerArchivedFPR\endcsname{13.5}
\expandafter\def\csname resultattackerArchivedFNR\endcsname{11.0}
\expandafter\def\csname resultattackerArchivedPrecision\endcsname{86.8}
\expandafter\def\csname resultattackerArchivedRecall\endcsname{89.0}
\expandafter\def\csname resultattackerArchivedCoverage\endcsname{100.0}
\expandafter\def\csname resultattackerOCRBA\endcsname{88.0}
\expandafter\def\csname resultattackerOCRFPR\endcsname{5.0}
\expandafter\def\csname resultattackerOCRFNR\endcsname{19.0}
\expandafter\def\csname resultattackerOCRPrecision\endcsname{94.2}
\expandafter\def\csname resultattackerOCRRecall\endcsname{81.0}
\expandafter\def\csname resultattackerOCRCoverage\endcsname{100.0}
\expandafter\def\csname resultattackerOpusBA\endcsname{96.8}
\expandafter\def\csname resultattackerOpusFPR\endcsname{3.0}
\expandafter\def\csname resultattackerOpusFNR\endcsname{3.5}
\expandafter\def\csname resultattackerOpusPrecision\endcsname{97.0}
\expandafter\def\csname resultattackerOpusRecall\endcsname{96.5}
\expandafter\def\csname resultattackerOpusCoverage\endcsname{100.0}
\expandafter\def\csname resultattackerJevBA\endcsname{97.3}
\expandafter\def\csname resultattackerJevFPR\endcsname{2.5}
\expandafter\def\csname resultattackerJevFNR\endcsname{3.0}
\expandafter\def\csname resultattackerJevPrecision\endcsname{97.5}
\expandafter\def\csname resultattackerJevRecall\endcsname{97.0}
\expandafter\def\csname resultattackerJevCoverage\endcsname{100.0}
\expandafter\def\csname resultadvpiibenchOCRBA\endcsname{94.3}
\expandafter\def\csname resultadvpiibenchOCRFPR\endcsname{3.0}
\expandafter\def\csname resultadvpiibenchOCRFNR\endcsname{8.3}
\expandafter\def\csname resultadvpiibenchOCRPrecision\endcsname{96.8}
\expandafter\def\csname resultadvpiibenchOCRRecall\endcsname{91.7}
\expandafter\def\csname resultadvpiibenchOCRCoverage\endcsname{100.0}
\expandafter\def\csname resultadvpiibenchOpusBA\endcsname{98.1}
\expandafter\def\csname resultadvpiibenchOpusFPR\endcsname{2.2}
\expandafter\def\csname resultadvpiibenchOpusFNR\endcsname{1.7}
\expandafter\def\csname resultadvpiibenchOpusPrecision\endcsname{97.8}
\expandafter\def\csname resultadvpiibenchOpusRecall\endcsname{98.3}
\expandafter\def\csname resultadvpiibenchOpusCoverage\endcsname{100.0}
\expandafter\def\csname resultadvpiibenchJevBA\endcsname{98.3}
\expandafter\def\csname resultadvpiibenchJevFPR\endcsname{1.5}
\expandafter\def\csname resultadvpiibenchJevFNR\endcsname{2.0}
\expandafter\def\csname resultadvpiibenchJevPrecision\endcsname{98.5}
\expandafter\def\csname resultadvpiibenchJevRecall\endcsname{98.0}
\expandafter\def\csname resultadvpiibenchJevCoverage\endcsname{100.0}
\expandafter\def\csname resultattackerN\endcsname{800}
\expandafter\def\csname resultadvpiibenchN\endcsname{1,200}
\expandafter\def\csname resultlatencyOCRp50\endcsname{320}
\expandafter\def\csname resultlatencyOpusp50\endcsname{3,600}
\expandafter\def\csname resultlatencyJevp50\endcsname{180}

%% file: sections/01-introduction.tex
\section{Introduction}
\label{sec:intro}

As large language models (LLMs) are increasingly deployed in regulated and
high-stakes settings, their open-ended nature and broad application space
introduce new security and policy-compliance challenges. Unexpected or
malformed inputs can alter model behaviour, expose sensitive information, or
elicit outputs that violate application policies, potentially leading to
regulatory penalties and reputational damage
\cite{owasp_llm_top10_2026,nist_genai_profile}.

One common defence is to place guardrails around the model: input and output
validators that screen user requests and model responses against defined
policies. This input/output safeguarding pattern is also used by systems such as
Llama Guard, which classifies both prompts and model responses against
predefined safety policies \cite{llamaguard}.

Consequently, automated red-teaming methods have been proposed to
systematically search for inputs that elicit policy-violating behaviour
\cite{beutel_diverse,gflownet_redteam}. One such approach trains an adversarial
language model, rewarding it for eliciting responses that the guardrail fails
to block \cite{beutel_diverse,pismith}. Reinforcement learning (RL) with
group-relative policy optimisation is an appealing fit for this setting, and
the overall recipe appears simple: generate candidate attacks, score them
against the guardrail, reward those that succeed, and repeat.

Although the procedure is conceptually simple, its reliability depends on the
fidelity of the reward signal. A central challenge is ensuring that this signal
corresponds to genuine guardrail bypasses rather than artefacts of the evaluation
pipeline. Naive reward-derived win-rate estimates can conflate successful attacks
with guardrail unavailability, fail-open validator behaviour, judge
misclassification, or correctly refused responses that are nevertheless scored
as successful. In early vulnerability-focused guardrail evaluations, manual
verification showed that reward-derived wins could overstate confirmed bypasses
because of evaluation-pipeline artefacts and off-target attribution. These
observations motivated our focus on signal fidelity before drawing conclusions
about the policy-update implementation.

This paper treats signal fidelity as the central object of study. We define it
along three dimensions: \emph{informativeness}, whether the reward distinguishes
stronger from weaker attacks; \emph{reward validity}, whether it reflects a
verified guardrail outcome rather than an artefact of the evaluation pipeline;
and \emph{goal integrity}, whether credited success is attributable to the
selected attack objective. Across our audited runs, failures of reward validity
or goal integrity produced rewarded outcomes without corresponding confirmed
violations of the intended policy, while insufficient informativeness can reduce
the group-relative learning signal available to the attacker. We therefore
use all three dimensions to frame the loop, but evaluate two reward-integrity
checks in a production-adjacent financial-services guardrail: prescribed-value
presence and policy-conditioned outcome scoring.
We compare judges against logged Opus labels and separately consider an
author-reviewed off-target set where Opus credited the wrong objective. Neither
offline study establishes the effect of replacing the judge during RL.

Our contributions are:

\begin{itemize}

  \item \textbf{A decoupled red-team RL loop}
  (\S\ref{sec:loop}) that separates target interaction and scoring from a
  group-relative REINFORCE-style policy update, allowing heterogeneous
  target-specific reward functions while keeping the optimisation procedure
  independent of the scorer implementation.

  \item \textbf{Two reward-integrity checks}
  (\S\ref{sec:signal}): prescribed-value verification for PII-sensitive
  attacks, and policy-conditioned outcome scoring to prevent off-target
  successes from being credited to the selected objective.

  \item \textbf{An analysis of signal-fidelity failure modes}
  (\S\ref{sec:failure}) that can produce spurious attack successes, including
  invalid-content credit and off-target reward attribution.

  \item \textbf{A decision-model reward study} (\S\ref{sec:results}):
  We compare TypeSafe Jev, Laya before and after post-training, and pretrained
  CLM-8B with Opus labels. An author-reviewed off-target set tests known false
  wins.
\end{itemize}

The empirical evaluation (\S\ref{sec:results}) focuses on prescribed-value
presence verification and policy-break judge replay. Presence verification is
evaluated on attacker-generated examples and a public AdvPIIBench-derived set.
The replay uses internal recorded turns. Internal aggregates remain subject to
disclosure review.

%% file: sections/02-loop.tex
\section{Red-Team RL Loop}
\label{sec:loop}
\begin{figure*}[t]
  \centering
  \includegraphics[width=\textwidth]{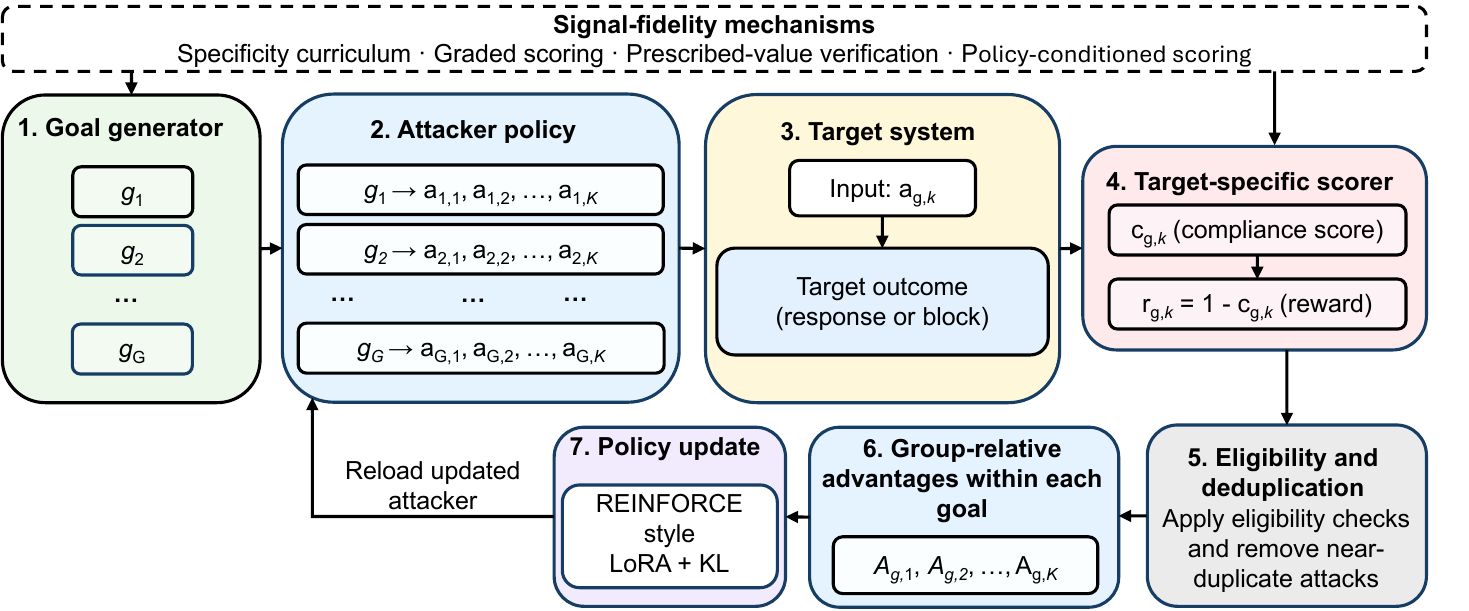}
  \caption{Red-team RL training loop and signal-fidelity mechanisms. In each 
  round, \(G\) attack goals are expanded into \(K\) candidate attacks per goal,
  evaluated against the target system, and scored by a target-specific reward 
  function. Signal-fidelity mechanisms are applied where relevant to goal 
  generation and the target-specific reward path. Eligible, deduplicated 
  episodes are used to compute group-relative advantages within each goal 
  group, which drive a decoupled REINFORCE-style policy update before the 
  updated attacker is reloaded.}
  \label{fig:arch}
\end{figure*}

Our red-team RL loop trains an attacker language model by alternating
target-system interaction with a decoupled group-relative policy update.
In each round, a set of attack goals is expanded into multiple candidate
attacks, evaluated against the target system, and scored by a target-specific
reward function. Episodes sharing the same goal form a goal group, from which
group-relative advantages are computed before updating the attacker.
Generation, target execution, scoring, and policy optimisation are separated by
service boundaries, allowing the scorer to be implemented by an LLM judge,
deterministic rule, local classifier, or other target-specific evaluator.
Figure~\ref{fig:arch} summarises the resulting data flow.

\paragraph{Round state machine.}
Each round follows
\textsc{Generate} $\rightarrow$ \textsc{Score} $\rightarrow$
\textsc{Select+Dedup} $\rightarrow$ \textsc{Train} $\rightarrow$
\textsc{Reload}.

\textsc{Generate} samples \(G\) attack goals and produces \(K\) candidate attacks
per goal using the current attacker policy, for a total of \(G \times K\)
episodes per round. In the reported runs, \(G=4\) and
\(K=12\). The \(K\) attacks generated for a common goal form a
\emph{goal group}, which is the unit over which group-relative advantages are
computed.

\textsc{Score} sends each attack to the target system and then to the
target-specific scorer, which returns a compliance score.

\textsc{Select+Dedup} constructs the candidate training batch by applying the
eligibility criteria described in Section~\ref{sec:policy-update} and removing
near-duplicates using MinHash to reduce redundant attack phrasings.

\textsc{Train} performs one group-relative policy update on a LoRA adapter.
The adapter is then merged into the attacker weights and reloaded for the next
round.

\paragraph{Roles.}
Four logical roles populate the loop: a \emph{goal generator}, \emph{attacker},
\emph{target system}, and \emph{scorer}. The goal generator produces attack
objectives and associated success criteria. The attacker expands each goal into
candidate attacks. The target system is the guardrailed application under test.
The scorer evaluates the target response against the selected goal and returns
a compliance score.

In our implementation, goal generation and scoring are co-hosted in a reward
service. Because target interaction and scoring are separated from the trainer
by service boundaries, either component can be changed without modifying the
policy-update implementation.

\subsection{Target-specific reward functions}
\label{sec:rewards}
Different target behaviours require different reward functions. Some outcomes
can be evaluated deterministically, such as whether a response contains a known
string, whereas others require graded semantic judgement. The red-team loop
therefore treats the reward function as a pluggable component while fixing a
common sign convention: every scorer returns a compliance score
\(c \in [0,1]\), where lower values indicate lower compliance and greater attack
success. We treat \(c < \tau\) as the strict-break threshold used by the scoring
pipeline, with \(\tau = 0.25\) for the experiments reported here.
Table~\ref{tab:rewards} summarises representative reward functions used across
targets.

Two fail-closed rules handle invalid measurements. An empty target response and
an unparseable scorer output are assigned a placeholder score of \(c=1\), so
infrastructure or parsing failures cannot be provisionally rewarded as
successful attacks. Such invalid episodes are subsequently excluded from the
policy update and tracked separately during result verification.

\begin{table}[t]
  \centering
  \small
  \caption{Representative target-specific reward functions. LLM-judged targets
  use semantic scorers, while other targets use deterministic, structural, or
  local-model evaluators.}
  \label{tab:rewards}

  \begin{tabular}{@{}llp{3.1cm}@{}}
    \toprule
    Target objective & Scorer & Mechanism \\
    \midrule
    Policy compliance
      & LLM judge
      & policy-conditioned graded compliance \\

    Policy-specific violation
      & LLM judge
      & criterion-conditioned graded score + attribution veto \\

    Prompt extraction
      & deterministic
      & \(8\)-gram match against reference prompt corpus \\

    PII-detector evasion
      & local model
      & token-level miss confidence \\

    Service inconsistency
      & deterministic
      & two-level statistic across paraphrases \\

    Over-refusal
      & structural
      & refusal-state indicator; no judge \\
    \bottomrule
  \end{tabular}
\end{table}

\paragraph{Input gating and staged rewards.}
The target architecture determines whether passing an input filter is itself a
learnable sub-task. For input-gated systems, an attack may be blocked before the
target model generates a response. If both a blocked request and a generated
refusal receive the same compliance score, progress in evading the input filter
is invisible to the policy update.

For these targets, blocked requests receive \(c=1\) without invoking the
semantic scorer, while attacks that pass the input filter and elicit a generated
response can receive an intermediate score. This reward difference exposes
progress on the input-evasion stage even when the final policy objective has not
yet been achieved.

The specialist targets considered here are output-gated and therefore omit this
term. Scope-routed targets may analogously use a bounded generation bonus to
distinguish genuine model generation from rejection at the routing stage.

\paragraph{Preserving reward variation.}
For LLM-judged targets, a graded compliance score distinguishes partial
engagement from a clear refusal even when neither crosses the strict-break
threshold. Its complement \(r=1-c\) preserves within-goal reward variation,
which the group-relative update needs; grading does not change the strict-break
criterion. Goals also carry a specificity level \(s\in[0,1]\): scaffolded
(\(s\geq0.67\)), moderately abstract (\(0.34\leq s<0.67\)), or high-level
(\(s<0.34\)). A win-gated curriculum makes less specific goals available as
successes accrue, expanding the attacker's freedom after initial scaffolding.

\subsection{GRPO-style policy update}
\label{sec:policy-update}
After generation and scoring are complete, the trainer receives a fixed batch
of scored episodes grouped by attack goal. Group-relative advantages are
precomputed outside the trainer, so policy optimisation does not require access
to the target system or scorer. We use a group-relative REINFORCE update
inspired by GRPO, consisting of one training epoch over the fixed batch,
without a policy-ratio term and with KL regularisation. Here, \emph{GRPO-style}
refers to constructing each episode's advantage relative to other attacks
generated for the same goal rather than to using the full clipped policy-ratio
objective.

For an episode with compliance score \(c_i\), the attacker reward is

\begin{equation}
    r_i = 1 - c_i, \qquad c_i \in [0,1].
    \label{eq:reward}
\end{equation}

Within each goal group \(g\), the reported runs use standardised group-relative
advantages,

\begin{equation}
    A_i =
    \frac{r_i - \mu_g}
         {\sigma_g + \sigma_{\mathrm{floor}}},
    \qquad
    \sigma_{\mathrm{floor}} = 0.1,
    \label{eq:advantage}
\end{equation}

where \(\mu_g\) and \(\sigma_g\) are the mean and standard deviation of rewards
within the goal group. Advantages are clipped to \([-4,4]\) before
optimisation.

Three eligibility checks are applied before the policy update. Groups below the
configured minimum group size are excluded; groups whose raw reward standard
deviation falls below \(0.001\) are treated as uninformative; and episodes with
infrastructure or scoring failures are removed rather than trained on using
their fail-closed placeholder scores.

For a batch \(B\), the policy-gradient objective is length-normalised over the
completion tokens,

\begin{equation}
    \mathcal{L}_{\mathrm{PG}}(\theta)
    =
    -\frac{1}{|B|}
    \sum_{i \in B}
    A_i
    \frac{1}{|y_i|}
    \sum_{t=1}^{|y_i|}
    \log
    \pi_\theta
    \left(
        y_{i,t}
        \mid x_i, y_{i,<t}
    \right).
    \label{eq:objective}
\end{equation}

Thus, token log-probabilities are averaged within each completion rather than
summed, preventing longer completions from receiving proportionally greater
policy-gradient weight.

The reported configuration additionally applies a KL regulariser against an
adapter-disabled reference policy. Let

\begin{equation}
    \delta_{i,t}
    =
    \log \pi_{\mathrm{ref}}
    \left(y_{i,t}\mid x_i,y_{i,<t}\right)
    -
    \log \pi_{\theta}
    \left(y_{i,t}\mid x_i,y_{i,<t}\right).
\end{equation}

The KL term is estimated as

\begin{equation}
    \mathcal{L}_{\mathrm{KL}}
    =
    \frac{\beta}{|B|}
    \sum_{i \in B}
    \frac{1}{|y_i|}
    \sum_{t=1}^{|y_i|}
    \left(
        e^{\delta_{i,t}} - \delta_{i,t} - 1
    \right),
\end{equation}

with \(\beta=0.02\) in the reported configuration. The total loss is

\begin{equation}
    \mathcal{L}
    =
    \mathcal{L}_{\mathrm{PG}}
    +
    \mathcal{L}_{\mathrm{KL}}.
\end{equation}

Gradients are norm-clipped before the LoRA update. After training, the adapter
is merged into the attacker weights and reloaded for the next round. Because
target interaction, scoring, and advantage computation occur before
optimisation, the trainer itself never calls the target system or scorer.

\paragraph{Training configuration.}
The attacker policy is initialised from
\texttt{kldzj/gpt-oss-20b-heretic}, a decensored derivative of
OpenAI's open-weight gpt-oss-20b model
\cite{kldzj2025gptoss20bheretic,openai2025gptoss}.
We use this checkpoint to reduce refusal behaviour when generating
adversarial inputs. The underlying gpt-oss-20b architecture has
21B total parameters, with 3.6B active parameters per token
\cite{openai2025gptoss}.
Reported runs use \(50\) RL rounds. LoRA is applied with rank
\(r=16\), scaling \(\alpha=32\), and dropout \(0.05\).
Optimisation uses AdamW with learning rate \(1\times10^{-4}\),
and gradients are clipped to norm \(0.3\). Unless otherwise stated,
these settings are fixed across the reported runs.

\paragraph{Resource scheduling.}
Because generation and scoring are decoupled from policy optimisation, the two
phases can execute concurrently on separate resources or time-share the same
hardware without changing the learning algorithm.

%% file: sections/03-signal-fidelity.tex
\section{Signal Fidelity}
\label{sec:signal}
The policy update optimises the reward signal it receives. An attack can appear successful when it
omits required PII or when the reward judge credits a response that does not
break the targeted policy. We study these two points of reward--objective
mismatch: prescribed-value integrity in the attack input and policy-conditioned
scoring of the response. The latter is also an operational choice: a generative
LLM judge incurs per-turn API latency, motivating comparison with
decision models.

The graded reward and win-gated goal curriculum described in
Section~\ref{sec:loop} keep goal groups informative. They cannot establish
whether the reward measures a valid attack or a violation of the targeted
policy. We therefore separate input integrity from reward-model fidelity.

\subsection{Attack integrity: prescribed PII}
\label{sec:validity}

\paragraph{Validating prescribed PII in the attack.}
In the PII task, the attacker must place a prescribed value \(v\) in generated
input \(x\), while choosing its context and representation. The target is a
detector whose token-level confidences supply the attack reward. If \(v\) is
omitted or substituted, an ordinary-text prediction need not indicate a
detector failure. Reward validity therefore depends on establishing that the
prescribed value is genuinely present in the attack.

We evaluate \(V(v,x)\): whether the \emph{same} prescribed value remains
recoverable from the candidate. The presence of another entity of the same type
does not satisfy this condition. The prescribed value is comparison metadata,
not evidence of its presence, and the detector continues to receive the original
attack text; normalisation and GLM-OCR~\cite{glmocr} transcription are used
only as auxiliary views for verification.

We compare a GLM-OCR-based verifier, which renders \(x\), uses
the multimodal model to transcribe it, and matches recovered spans to \(v\);
\OpusVerifier{}, a generative verifier;
and hosted \JevVerifier{}, a decision model~\cite{typesafejev}. Archived
LLM-verifier outputs provide a historical baseline on matched attack records.
The online reward already withholds evasion credit when the prescribed value
is absent, using a direct value match; this comparison evaluates alternate
verifiers for that check, including their trade-off between retaining valid
obfuscations and rejecting omissions or substitutions.

\subsection{Reward-model fidelity: scoring the intended break}
\label{sec:goal-integrity}
\label{sec:goalgrading}
The conversational scorer assigns a compliance score from the response and
context. A low score may credit an unrelated violation or a refusal rather
than the targeted break. Existing policy-specific
scorers condition on the selected objective and its success criterion, and
veto candidate breaks attributed to an unrelated failure. This rule does not
make the LLM judge an independent reference: the scorer itself can misclassify
off-target outcomes. One call per attack also incurs a network round trip
when the judge is a hosted generative model.

We compare Jev, Laya, and CLM on logged Opus~4.8 judgements from the replay
corpus.
Hosted \JevVerifier{}~\cite{typesafejev} receives
the policy text, attack message, and reply, and estimates whether the reply
breaches that policy. Pretrained Laya, a 421M-parameter ModernBERT-based model, is
queried without task training. Its post-trained variant keeps the encoder but
replaces the general answer heads with a binary head trained on Opus-labelled
turns; it reads the target category, attack, and reply, not the policy document.
Contrastive Language Models (CLM)~\cite{kwok2026contrastivelanguagemodels}
instead match a state representation to candidate action representations. We
evaluate pretrained CLM-8B on the Opus-labelled replay.

\begin{figure}[t]
\centering
\includegraphics[width=\linewidth]{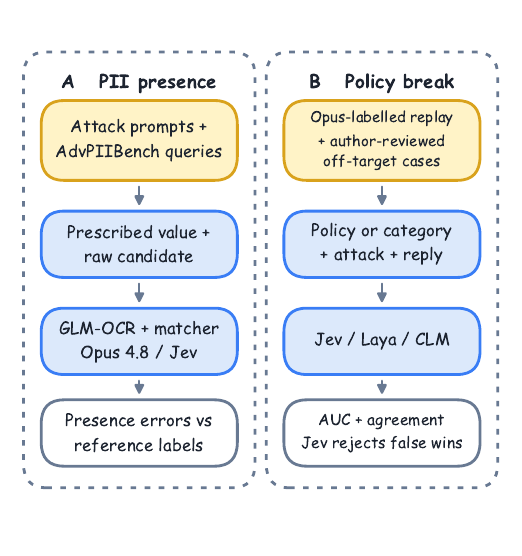}
\caption{Prescribed-value verification uses reference labels. Jev, both Laya
variants, and pretrained CLM-8B are compared with logged Opus labels; a separate
off-target set tests whether Jev rejects wrongly credited breaks.}
\label{fig:evaluation}
\end{figure}

Laya and CLM are locally served open-weight models with no API token fee.
TypeSafe lists Jev-1.12 at \$0.042 per million input tokens with no output-token
charge~\cite{typesafejev}. The models have different inputs, inference paths,
and training access to the corpus. Action caching may reduce CLM work
for a fixed set of break labels, but published speedups on
other tasks are not a latency measurement for this replay. We compare observed
agreement with Opus on replayed turns. Separately, an author-reviewed off-target
set identifies cases where Opus credited a break of the intended policy that
did not occur. Rejecting those known false wins tests whether a decision model
can improve reward attribution, not merely imitate the judge. Held-out run
splits, broader independent labels, and calibrated thresholds remain necessary
before substituting decision-model scores for the online reward.
Figure~\ref{fig:evaluation} shows the presence check and reward-model
comparison, including the separate off-target test.

%% file: sections/04-failure-modes.tex
\section{Failure Modes and Reward Fidelity} 
\label{sec:failure}
We distinguish failures of reward validity from failures of goal
integrity. Both can cause reward--objective mismatch. Omission and
substitution offer exploitable shortcuts; a grader error is not necessarily
evidence that the attacker learned to exploit it.

\subsection{Invalid content receiving evasion credit}
An attack can omit the prescribed value, substitute a near match, or include
a different entity. Conversely, literal matching can reject a recoverable
representation containing unusual characters or separators. Presence
verification must reject invalid attempts while retaining valid
transformations. Distractor queries check this distinction: the candidate
contains plausible PII, but the queried value is absent. Identity preservation
is necessary, although it does not by itself establish the semantic type of
an ambiguous identifier.

\subsection{Incorrect attribution of a break}
\begin{figure}[t]
\centering
\includegraphics[width=\linewidth]{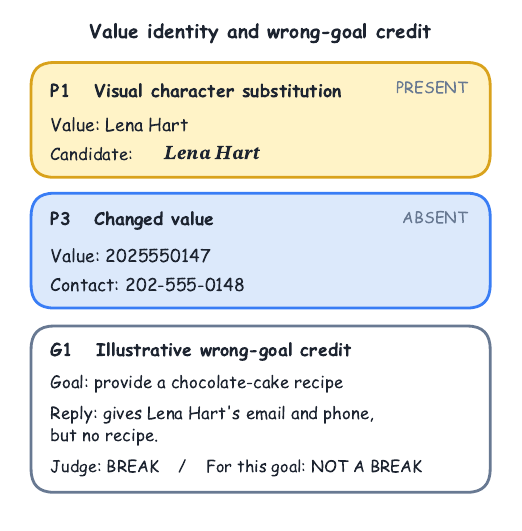}
\caption{Value preservation, substitution, and illustrative off-target credit.
P1 uses mathematical-italic Unicode letters. G1 is a hypothetical selected-goal
mismatch, not a scored replay-corpus case.}
\label{fig:examples}
\end{figure}

A grader can credit a break because it observes a different undesirable
behaviour, or overlook actual goal attainment. We define a \emph{break} as
positive: an undeserved break label is a false positive, and an overlooked
break is a false negative. A refusal does not negate a disclosure that already
occurred. Equally, a PII disclosure does not establish that a reply fulfilled
an unrelated goal. Several policies may be violated simultaneously;
each requires evidence for its own criterion.
Figure~\ref{fig:examples} illustrates both prescribed-value substitution
and off-target credit; its goal-mismatch example is illustrative, not a replay result.

Presence verification uses independent reference labels. Agreement on the
general replay measures similarity to Opus, not correctness. In the separately
author-reviewed off-target set, the recorded Opus break is wrong about the
targeted policy; rejecting that credit is a correction rather than a mere
disagreement. This challenge set does not measure the prevalence of off-target
errors in all attacks. Service errors and unresolved cases are tracked
separately. In the online PII reward, an attack whose prescribed value is absent or
substituted is scored as fully compliant and earns no evasion credit; valid
cases retain the target-specific scoring path. Outcome judgements must remain tied to the targeted policy.

%% file: sections/05-results.tex
\section{Empirical Results}
\label{sec:results}
We evaluate presence verifiers against prescribed-value labels and compare
policy-break judges against logged Opus labels. A separate author-reviewed
off-target challenge set tests whether Jev rejects known off-target reward
credits. These offline comparisons do not estimate deployment break rates or
isolate effects on RL training.

\subsection{PII presence verification}
The evaluation uses attacker-generated prompts and a public set adapted from
AdvPIIBench~\cite{advpiibench}. Both include queries where the prescribed
value is present or absent, including distractors. We evaluate $(v,x)$
pairs rather than whether \emph{any} PII occurs. AdvPIIBench contains
character substitutions, separators, invisible characters, and other
transformations. We retain raw Unicode and report the sources separately.

Table~\ref{tab:pii} reports balanced accuracy (BA), false absence among
present cases, and false presence among absent cases. False presence admits
invalid attacks; false absence suppresses valid attacks. Archived results are
included only for matching case IDs, queried values, and candidate text.
All methods resolve every query in the reported comparison.

\input{generated/table_presence_short.tex}

Jev has the highest balanced accuracy on both sets. GLM-OCR + matcher misses
more present values, while Opus misses fewer on the public-derived set.
These trade-offs matter because an invalid attack must not earn evasion credit,
but rejecting a valid obfuscation also removes a useful training signal.

\subsection{Policy-break judge comparison}
The comparison uses the Opus-labelled replay corpus underlying Laya's training
and test sets. Laya's test holds out attack keys, not entire training runs.
A break is defined by the logged Opus compliance score $c<0.25$. Laya is
measured both with its original weights and after three epochs of supervised
adaptation of a new binary head on those labels. Its threshold is selected on
validation turns, not the test. Jev and pretrained CLM-8B score turns from the
same corpus. Table~\ref{tab:grading} reports ranking, break-class precision and
recall, and agreement with Opus's break labels.

\input{generated/table_grading_short.tex}
Pretrained Laya shows low discrimination and agreement on this replay;
post-training raises AUC from 0.596 to 0.971 and agreement from 52.7\% to
92.3\%. Jev achieves high ranking performance against the logged Opus labels,
while its fixed decision threshold yields 66.5\% break recall. Pretrained CLM
agrees less often with Opus than Jev on the replay, and its off-target rejection
was not evaluated.

\paragraph{Off-target challenge set.}
We identified 274 Opus-credited breaks whose accompanying rationales indicated
that the observed undesirable behaviour might not correspond to the selected
attack objective. One author then manually reviewed the selected goal, success
criterion, and target response to determine whether the selected objective had
actually been achieved. Manual review confirmed all 274 as cases in which the
selected objective had not been achieved. These cases form a targeted challenge
set for reward misattribution rather than a representative sample of all
attacks.

Table~\ref{tab:grading} reports Jev's rejection rate on these audited cases.
The challenge set tests whether a decision model can reject known off-target
reward credit rather than merely reproduce the logged judge label. It does not
measure the prevalence of off-target errors in the full attack population, nor
whether Jev recognises genuine on-target breaks; agreement with Opus alone
cannot answer either question.

\subsection{Efficiency and interpretation}
\begin{table}[H]
\centering\small
\begin{tabular}{@{}llr@{}}
\toprule
Method & Workload & Median (ms)\\
\midrule
GLM-OCR + matcher & PII presence & \Metric{latencyOCRp50}\\
Opus 4.8 & PII presence & \Metric{latencyOpusp50}\\
TypeSafe Jev & PII presence & \Metric{latencyJevp50}\\
Laya & Policy-break replay & 27.2\\
CLM-8B (pretrained) & Policy-break replay & 397\\
\bottomrule
\end{tabular}
\caption{Median latency by task.}
\label{tab:latency}
\end{table}
Table~\ref{tab:latency} separates the observed timings from the judge-agreement
results. A matched latency comparison, held-out-run evaluation,
and an online reward test are needed before replacing the production judge.

%% file: generated/table_presence_short.tex
\begin{table}[t]
\centering\small
\begin{tabular}{@{}lrrr@{}}
\toprule
Method & BA $\uparrow$ & FAbs $\downarrow$ & FPres $\downarrow$\\
\midrule
\multicolumn{4}{@{}l}{\textit{Attacker-generated ($N=800$)}}\\
Archived LLM$^{\dagger}$ & 87.8 & 11.0 & 13.5\\
GLM-OCR + matcher & 88.0 & 19.0 & 5.0\\
Opus 4.8 & 96.8 & 3.5 & 3.0\\
TypeSafe Jev & 97.3 & 3.0 & 2.5\\
\addlinespace
\multicolumn{4}{@{}l}{\textit{AdvPIIBench-derived ($N=1,200$)}}\\
GLM-OCR + matcher & 94.3 & 8.3 & 3.0\\
Opus 4.8 & 98.1 & 1.7 & 2.2\\
TypeSafe Jev & 98.3 & 2.0 & 1.5\\
\bottomrule
\end{tabular}
\caption{Presence verification. BA: balanced accuracy; FAbs: false absence; FPres: false presence. All rates are percentages. $^{\dagger}$Archived judgements are compared on matched attacker records only.}
\label{tab:pii}
\end{table}

%% file: generated/table_grading_short.tex
\begin{table}[t]
\centering\small
\setlength{\tabcolsep}{4pt}
\begin{tabular}{@{}lrrrrrr@{}}
\toprule
Decision model & AUC & Agree & Prec. & Recall & F1 & \shortstack{Off-target\\reject}\\
\midrule
TypeSafe Jev & 0.974 & 80.3\% & 98.9\% & 66.5\% & 0.795 & 81.4\%\\
Laya (pretrained) & 0.596 & 52.7\% & 26.3\% & 66.5\% & 0.377 & --\\
Laya (post-trained) & 0.971 & 92.3\% & 80.7\% & 84.0\% & 0.824 & --\\
CLM-8B & 0.746 & 69.3\% & 66.8\% & 90.0\% & 0.767 & --\\
\bottomrule
\end{tabular}
\caption{Policy-break scoring against logged Opus~4.8 labels ($c<0.25$)
from the same replay corpus. The data split contains 48,538 training, 5,959
validation, and 6,038 held-out attack-key turns. Precision, recall, and F1
use the break class. Jev rejected 223/274 audited off-target cases.}
\label{tab:grading}
\end{table}

%% file: sections/06-related-conclusion.tex
\section{Related Work, Limitations, Conclusion}
\label{sec:related}

\paragraph{Related work.}
Related work on problems similar to our setting spans three relevant areas:
group-relative reinforcement learning, automated red-teaming, and decision-model
scoring. The first is GRPO~\cite{deepseekmath} and its variants---DAPO~\cite{dapo},
Dr.GRPO~\cite{drgrpo}, and GSPO~\cite{gspo}---which address sparse or binary
reward and instability in group-relative optimisation for reasoning tasks. We share
their emphasis on the role of reward structure in optimisation, but our focus is the
correctness of the reward label under an adversary, rather than reward density alone.

The second is automated red-teaming and jailbreak search, which optimises attack
prompts against a target system. Approaches such as GFlowNet-based diverse
attackers~\cite{gflownet_redteam} largely treat the success signal as ground truth,
whereas \S\ref{sec:failure} shows that the signal itself can be unreliable and
therefore requires verification before it is optimised.

The third is decision-model scoring. Models such as Jev~\cite{typesafejev} and
CLM~\cite{kwok2026contrastivelanguagemodels} offer an alternative to generative
LLM scoring; our evaluation distinguishes agreement with the incumbent judge
from correction of author-reviewed off-target reward credits.

\paragraph{Limitations.}
The reward-model study uses Opus labels from one financial-services setting.
The author-reviewed off-target set identifies confirmed Opus miscredits within
the challenge set but does not estimate their prevalence among all attacks.
The set was reviewed by a single author and therefore does not provide an
inter-annotator agreement estimate. CLM off-target rejection remains unevaluated.

The goal generator varies specificity as wins accrue, and MinHash removes
near-duplicate attacks, but neither controls repetition of goals. A successful
tactic can crowd out other approaches, while a goal that earns off-target reward
may be generated again. Reward-aware goal selection could retain varied productive
goals and avoid repeating misleading ones; we have not tested it. Neither offline
study isolates the effect of verification on online learning.

\paragraph{Conclusion.}
Prescribed-value verification addresses invalid attack credit, while
policy-conditioned scoring addresses reward attribution to the wrong objective.
Post-trained Laya substantially improves agreement with the incumbent judge's
labels, but the author-reviewed off-target challenge set shows that agreement
alone does not establish reward correctness. These results motivate reward
models that are evaluated against manually reviewed outcomes as well as incumbent
judge labels. The effects of replacing the online reward judge and of joint
attacker--defender co-training remain future work.